\documentclass[journal]{IEEEtran}

\usepackage{setspace}
\usepackage{multirow}
\usepackage{rotating}

\usepackage[finalnew]{trackchanges} 

\usepackage{cite}
\usepackage{hyperref}
\hypersetup{
    colorlinks,
    citecolor=black,
    filecolor=black,
    linkcolor=black,
    urlcolor=black
}
\usepackage[nameinlink, capitalize]{cleveref}

   \usepackage{graphicx}

\begin{document}
%
\title{Next Day Wildfire Spread:\\ A Machine Learning Data Set to Predict Wildfire Spreading from Remote-Sensing Data}
%
%
%

\author{Fantine~Huot\textsuperscript{1}\thanks{\textsuperscript{1}fantine@sep.stanford.edu},
        R.~Lily~Hu\textsuperscript{2}\thanks{\textsuperscript{2}rlhu@google.com},
        Nita~Goyal,
        Tharun~Sankar,
        Matthias~Ihme,
        and~Yi-Fan~Chen
        }

\maketitle


\begin{abstract}
Predicting wildfire spread is critical for land management and disaster preparedness. To this end, we present `Next Day Wildfire Spread,' a curated, large-scale, multivariate data set of historical wildfires aggregating nearly a decade of remote-sensing data across the United States. In contrast to existing fire data sets based on Earth observation satellites, our data set combines 2D fire data with multiple explanatory variables (e.g., topography, vegetation, weather, drought index, population density) aligned over 2D regions, providing a feature-rich data set for machine learning.  To demonstrate the usefulness of this data set, we implement a \add{neural network} that takes advantage of the spatial information of this data to predict wildfire spread. We compare the performance of the neural network with other machine learning models: logistic regression and random forest. This data set can be used as a benchmark for developing wildfire propagation models based on remote sensing data for a lead time of one day.
\end{abstract}


%
\IEEEpeerreviewmaketitle

\section{Introduction}

\add{Wildfires can threaten} livelihood and properties, and impact environment and health \cite{vanderwerfglobalfire,kollanusmortality}. Over the last few decades, wildfire management has changed profoundly, facing longer fire seasons and more severe fires with more acres burned on average each year \cite{westerling2006warming}. In 2019, more than 775,000 residences across the United States were flagged as at ``extreme'' risk of destructive wildfire, amounting to an estimated reconstruction cost of 220 billion dollars \cite{corelogicwildfire}. Wildfires significantly impact the climate and are estimated to contribute to 10\% of the CO$_2$ emissions per year worldwide~\cite{vanderwerfglobalfire}. Furthermore, the  health impact due to wildfire aerosols is estimated at 300,000 premature deaths per year~\cite{kollanusmortality}.

There is a \add{need} for novel wildfire warning and prediction technologies that enable better fire management, mitigation, and evacuation decisions. In particular, evaluating the fire likelihood \cite{brillinger2003risk} -- the probability of wildfire burning in a specific location -- would provide valuable land management and disaster preparedness capabilities. Predictions of where a fire will spread in the upcoming day are essential to wildfire emergency response, leading to optimal allocation of resources and quick response to fire activity \cite{thompson2016application,thompson2019risk}. Additionally, identifying regions of high wildfire likelihood offers the possibility for targeted fire-prevention measures, such as shutting down power lines to minimize electrical fire hazards \cite{jazebi2019review,sharma2020pin}, or compartmentalizing forests by creating vegetation fuel breaks \cite{rigolot2002fuel,marino2014forest}.

The increase in the availability of remote-sensing data, computational resources, and advances in machine learning provide unprecedented opportunities for data-driven approaches to estimate wildfire likelihood. 
Herein, we investigate the potential of deep learning models to predict wildfire spreading from observational data~\cite{sullivan2009wildland1,sullivan2009wildland2,sullivan2009wildland3,jain2020review}. For this purpose, we present a new data set: `Next Day Wildfire Spread.'

This `Next Day Wildfire Spread' data set is a curated, large-scale, multivariate data set of historical wildfires over nearly a decade across the United States, aggregated using Google Earth Engine (GEE)~\cite{gorelick2017google}. This data set takes advantage of the increasing availability and capability of remote-sensing technologies, resulting in extensive spatial and temporal coverage from a wide range of sensors (e.g., Moderate Resolution Imaging Spectroradiometer (MODIS)~\cite{modis}, Visible Infrared Imaging Radiometer Suite (VIIRS)~\cite{viirs}, Shuttle Radar Topography Mission (SRTM)~\cite{srtm}). Our data set combines historical wildfires with 11 observational variables overlaid over 2D regions at 1 km resolution: elevation, wind direction, wind speed, minimum temperature, maximum temperature, humidity, precipitation, drought index, vegetation, population density, and energy release component. The resulting data set has 18,545 fire events, for which we provide two snapshots of the fire spreading pattern, at time $t$ and $t + 1$~\textit{day}.

Among existing fire data sets based on Earth observation satellites \cite{laurent2018fry,andela2019global,artes2019global}, most do not include this many variables at 1 km resolution. \add{Fire data sets such as FRY} \cite{laurent2018fry}, \add{Fire Atlas} \cite{andela2019global}, \add{and the 1.88 million US Wildfires catalog} \cite{short2017spatial} focus on the total burn area \add{and do not provide the temporal resolution or the 2D information required for characterizing fire spreading patterns}. In contrast, GlobFire \cite{artes2019global} is designed for characterizing fires,
and includes daily timestamps of the fire perimeter. 
However, none of these data sets include other variables such as vegetation, weather, drought, or topography data \add{required for fire prediction}. \add{It is true that} previous studies \add{have} combined fire data with other variables \cite{diaz2001space,krawchuk2006biotic,verdu2012multivariate,ganteaume2013review,vecin2016biophysical,bui2017hybrid}, but these data sets \add{have not been} publicly released or only partially so. A data set of fire events in Canada \cite{sayad2019predictive} \add{has been released but is limited} in geographical coverage and the number \add{of recorded fire events (386 events)}. Table \ref{table:fire_data} compiles a summary of these publicly available wildfire data sets.

Benchmark data sets have played an essential role in driving progress and innovation in machine learning research~\cite{russakovsky2015imagenet,bennett2007netflix}. They enable performance comparisons across models, which in turn can lead to better predictive models. We make the `Next Day Wildfire Spread' data set available to the broader scientific community for benchmarks, model comparisons, and further insights for wildfire predictions\footnote{https://www.kaggle.com/fantineh/next-day-wildfire-spread}.  We share our GEE data aggregation code for users to adapt to their use cases\footnote{https://github.com/google-research/google-research/tree/master/simulation\_research/next\_day\_wildfire\_spread}. 

\begin{sidewaystable*}[!h]
\renewcommand{\arraystretch}{1.6}
\caption{Publicly Available Fire Data sets}
\label{table:fire_data}
\begin{center}
\begin{tabular}{l|cccccc}
    \hline
    Name & Fire information & Coverage & Period & Spatial resolution & Fire spreading & Other variables \\
    \hline
    FRY~\cite{laurent2018fry} & Total burn area & Worldwide & 2005 - 2011 & 500 m & N/A$^*$ & - \\
    Fire Atlas~\cite{andela2019global} & Total burn area & Worldwide & 2003 - 2016 & 500 m & N/A$^*$ & - \\
    GlobFire~\cite{artes2019global} & Active fire & Worldwide & 2001 - 2017 & 500 m & Daily fire maps & - \\
    1.88 million US Wildfires~\cite{short2017spatial} & Active fire & USA & 1992 - 2015 & Point coordinates & N/A$^*$ & - \\
    Fire events in Canada~\cite{sayad2019predictive} & Active fire & Regions in Canada & August 2014 & Point coordinates & N/A$^*$ & Vegetation, Surface Temperature \\
    \hline
     \multirow{4}{*}{Next Day Wildfire Spread} & \multirow{4}{*}{Active fire} & \multirow{4}{*}{USA} & \multirow{4}{*}{2012 - 2020} & \multirow{4}{*}{1 km} &  & Elevation, Wind Direction and Speed, \\ 
& & & & & $t$ and $t + 1$~\textit{day}& Min. and Max. Temperatures,  \\     
& & & & & fire maps & Humidity, Precipitation, Drought Index, \\
& & & & & & Vegetation, Population density, ERC \\
    \hline
\end{tabular}
\end{center}

\vspace{0.5cm}

\footnotesize{$^*$The total burn area by itself does not provide information about the fire spreading pattern.}
\end{sidewaystable*}

\add{In recent years, deep learning approaches have been increasingly adopted for prediction tasks from remote sensing data, as reviewed} in~\cite{jain2020review}. For instance, \cite{alonso2003intelligent} used a deep learning model to estimate the daily fire likelihood from temperature, humidity, and rainfall data from five weather stations in the Galicia region of Spain. On Lebanon data, \cite{sakr2011efficient} implemented a neural network for predicting the occurrence of wildfires based only on monthly relative humidity and cumulative precipitation data. \add{Compared to these two data sets, the `Next Day Wildfire Spread' data set has more data samples and a much greater geographical coverage.} In an Australian study, \cite{dutta2013deep} use neural networks on aggregated monthly meteorological data (evaporation, precipitation, incoming solar irradiance, maximum temperature, soil moisture, wind speed, pressure and humidity) to estimate the risk of fire occurrence. \add{However, with monthly timestamps, their data set does not have the temporal resolution for estimating fire spreading}.

For fire spread prediction, \cite{hodges2019wildland} use a convolutional neural network to predict spreading patterns from environmental variables such as topography and weather, but only use \add{synthetic} data from computational models. \add{In} \cite{radke2019firecast}, \add{a similar method is used} to predict fire spread but from remote-sensing data collected from geographic information systems (GIS), combining topography and weather (pressure, temperature, dew point, wind direction, wind speed, precipitation, and relative humidity). \add{The objectives of this study are similar to the ones in our work but differs in the methodology and data aggregation. The data set was limited to regions in Colorado and constrained to the data available in the Geospatial Multi-Agency Coordination (GeoMAC) system} \cite{walters2011geospatial}. \add{Moreover, the data sets from studies \cite{alonso2003intelligent,sakr2011efficient,dutta2013deep,hodges2019wildland,radke2019firecast} were not publicly released.}

\add{In contrast, our `Next Day Wildfire Spread' is an extensive publicly-available data set for ML fire spread prediction. Aggregating large data sets with many observational variables is challenging because it is usually limited by the data availability across multiple data sources. We tackle this challenge by using the GEE framework, which allows to capture a large ensemble of fire events and more observational variables than any of the aforementioned data sets, all at 1 km resolution instead of point-wise locations.  In addition, it is the only data set of this type to have a variable to represent anthropogenic activity. Moreover, its spatial dimension makes it suitable for computer vision and segmentation tasks.} 
\add{By releasing the data set with its interactive GEE aggregation code, we offer an end-to-end framework for taking advantage of GEE capabilities for developing ML data sets. We include the full preprocessing and data interface to make it as accessible as possible to ML practitioners.} 

To illustrate the usefulness of our `Next Day Wildfire Spread' data set, we train a deep learning model for fire spread prediction. Given a wildfire, we predict where the fire will spread the following day. We implement a convolutional autoencoder, \add{a specialized type of neural network}, to take advantage of the 2D information of this data set and compare its performance with two other ML models, namely logistic regression and random forest.

The remainder of this manuscript has the following structure.~\Cref{sec:data} describes our data aggregation workflow for historical remote-sensing data. In~\cref{sec:data_processing}, we discuss our processing pipeline for the fire spread predictions.~\Cref{sec:model} presents the deep learning model, and results are presented in~\cref{sec:ongoing}. The article finishes with conclusions in~\cref{SEC_CONCL}.

\section{Data Set Aggregation}
\label{sec:data}
With the continuing improvements in quality, resolution, and coverage of remote sensing technologies, we now have access to increasingly more data for characterizing wildfires. Using data sources available in Google Earth Engine (GEE)~\cite{gorelick2017google}, we present a data aggregation workflow for creating the `Next Day Wildfire Spread' data set, combining historical fire events with remote sensing data.

\subsection{Motivation}
We collect the data at different locations and times at which wildfires occurred. We extract the data as 64$\,$km$\,\times\,$64$\,$km regions at 1~km resolution to capture all typical active fire sizes \cite{schoenberg2003distribution,malamud2005characterizing,dennison2014large}. We process the data from GEE to represent the fire information as a fire mask over each region, showing the locations of `fire' versus `no fire', with an additional class for uncertain labels (i.e., cloud coverage or other unprocessed data). We include both the fire mask at time $t$ and at time  $t + 1$~\textit{day} to provide two snapshots of the fire spreading pattern.

\begin{figure*}[!htb!]
  \centering
  \includegraphics[width=\textwidth,clip,trim={0cm 0.3cm 0cm 0cm}]{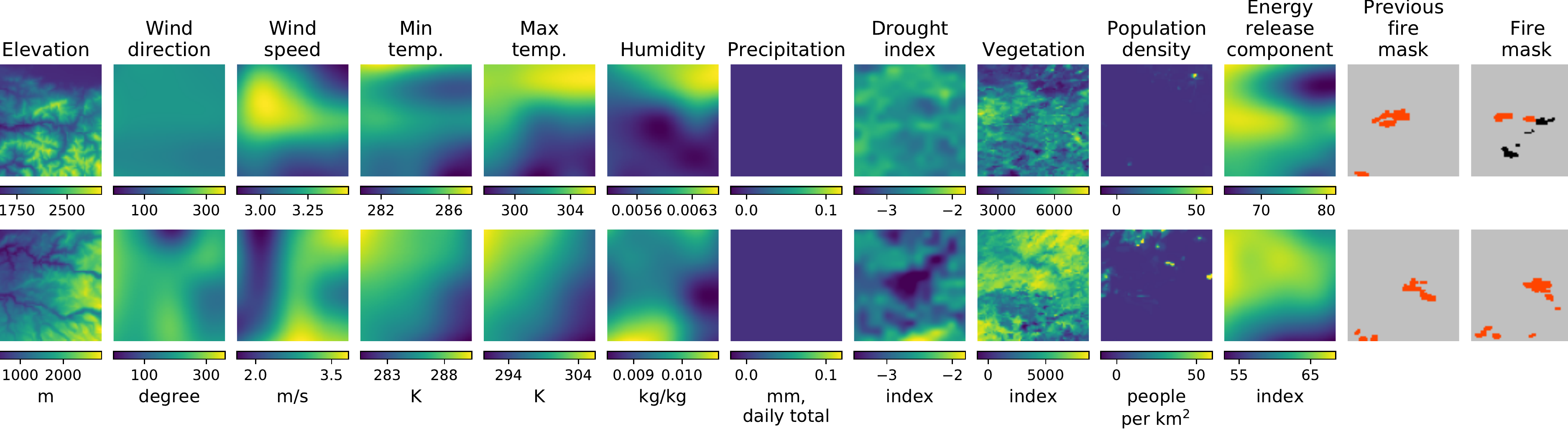}
  \caption{Examples from our `Next Day Wildfire Spread' data set. Each example is 64$\,$km$\,\times\,$64$\,$km at 1 km resolution. Each row corresponds to a location and time $t$ at which a fire occurred, and the columns represent the different variables. In each row, all the weather, drought, vegetation and population density variables are extracted at that same location at time $t$. `temp' denotes temperature, `wind direction' is the azimuth in degrees clockwise from North.  The `previous fire mask' corresponds to the fire locations at time $t$, while `fire mask' corresponds to the following day at $t + 1$~\textit{day}.  In the fire masks, red corresponds to fire, while grey corresponds to no fire. Black indicates uncertain labels (i.e., cloud coverage or other unprocessed data).}
  \label{fig:data_visualization}

\end{figure*}

Using GEE, we aggregate data from different data sources and align the data in location and time using the same projection (WGS84).
With this methodology, we combine these fire masks with variables that are of direct relevance to wildfire predictions \cite{stojanova2012estimating,vecin2016biophysical,van2018fire,dutta2013deep}: elevation, wind direction and wind speed, minimum and maximum temperatures, humidity, precipitation, drought index, normalized difference vegetation index (NDVI), and energy release component (ERC). In particular, NDVI, drought index, and the weather variables provide information relevant to fuel properties. \add{ERC is a calculated output of the National Fire Danger Rating System (NFDRS)} \cite{bradshaw19841978}. \add{It is considered a composite fuel moisture index as it reflects the contribution of all fuels to potential fire intensity}. In addition, we include population density to correlate the wildfires to anthropogenic ignitions and for risk-severity assessment. Humans cause 84\% of fires~\cite{Balch2946}, so we use population density as a proxy for anthropogenic activity. 
Examples from this data set are illustrated in~\cref{fig:data_visualization}.

In collecting this data set, we prioritized data from publicly available data sources that are regularly updated with recent data. This enables the methodologies developed on this data set to translate to future data predictions, including near real-time prediction. The use of openly available data with generous terms of use allows it to be incorporated into dynamic wildfire prevention and management tools for future applications with minimum licensing constraints.

This feature-rich data set has multiple applications. It can be used for statistical analysis or to study correlations between various variables. To demonstrate this capability, we use this data set for predicting wildfire propagation. The data set can be resampled to different region sizes, spatial resolutions, or any subset of the provided data. In addition to the data itself, we also release the GEE export code that generates the data set, giving users the flexibility of adapting it to their use cases\textsuperscript{2}. The export code can be changed to incorporate additional data sources, on a different geographic region or worldwide, over different time scales and intervals (hourly, daily, or weekly). For instance, \cite{huot2020deep} presents examples of the same data set extracted as weekly time sequences to create maps of the total burn area. 

\subsection{Data Sources from Earth Engine}
\label{sec:data_features}

We compile this data set from multiple remote-sensing data sources from GEE~\cite{gorelick2017google}. We selected data sources with extensive geographical and historical coverage, and reduced missing data:

\begin{itemize}
    \item \textbf{Historical wildfire} data are from the MOD14A1 V6 data set~\cite{modis}, a collection of daily fire mask composites at 1~km resolution since 2000, provided by NASA LP DAAC at the USGS EROS Center.
        
    \item \textbf{Topography} data are from the Shuttle Radar Topography Mission (SRTM)~\cite{srtm}, sampled at 30~m resolution.
        
    \item USA \textbf{Weather} data  are from the University of Idaho Gridded Surface Meteorological Dataset (GRIDMET)~\cite{abatzoglou2013development}, a collection of daily surface fields of temperature, precipitation, winds, and humidity at 4~km resolution since 1979, provided by the University of California Merced.
    
    \item USA \textbf{Drought} data is from GRIDMET Drought, a collection of drought indices derived from the GRIDMET data set~\cite{abatzoglou2014seasonal}, sampled at 4~km resolution every 5 days since 1979, provided by the University of California Merced.
        
    \item \textbf{Vegetation} data are from the Suomi National Polar-Orbiting Partnership (S-NPP) NASA VIIRS Vegetation Indices (VNP13A1) data set~\cite{viirs}, a collection of vegetation indices sampled at 0.5~km resolution every 8 days since 2012, provided by NASA LP DAAC at the USGS EROS Center.
    
    \item \textbf{Population density} data are from the Gridded Population of World Version 4 (GPWv4) data set by the Center for International Earth Science Information Network (CIESIN) \cite{gpw4}. This data contains the distribution of the global human population, sampled every 5 years at 1~km resolution.
\end{itemize}

All the variables in our data set are available from the data sources cited; none were calculated by us.

While we selected these data sources for creating our data set, users can change the data sources based on their needs using the provided GEE data export code. For instance, other fire mask data sets include FIRMS (Fire Information for Resource Management System) \cite{davies2008fire}, or GOES-16 and GOES-17 \cite{schroeder2010early, griffith2017closer}. Additional variables can be extracted and combined from the data from the MODIS Aqua and Terra satellites \cite{savtchenko2004terra}. As such, the data export code can be adapted to encompass a variety of wildfire problems beyond fire spread prediction, such as identifying fire precursors, probability of ignition, and long-term fire risk patterns.

\subsection{Data Aggregation}

From  these data sources,  we  sample examples at all locations and times with active fires. We consider fires separated by more than 10 km as belonging to a different fire. We extract both the fire mask at time $t$, which we denote `previous fire mask', and at time $t + 1$~\textit{day}, which we denote `fire mask'. For characterizing fire spreading, we only keep samples for which there was at least one area on fire within the region at time $t$; in other words, for which the `previous fire mask' contains any fire at all. 

We extract the data over the contiguous United States from 2012 to 2020. We select this time period due to data availability. We split the data between training, evaluating, and testing by randomly separating all the weeks between 2012 and 2020 according to an 8:1:1 ratio, respectively, while keeping a one-day buffer between weeks from which we do not sample data. 

Since the data sources have different spatial resolution, we align all the data to 1~km resolution, which corresponds to the spatial resolution of the fire masks. We downsample the topography and vegetation data and use bi-cubic interpolation for the weather and drought data. The data sources also have different temporal resolution and are refreshed over different time intervals. For variables that vary slowly over time, we take the last available time stamp at time $t$. For the weather data, refreshed 4 times a day, we take the average of each weather variable over the day corresponding to time $t$.

Using these data sources, we see in~\cref{fig:data_visualization} that variables such as elevation, vegetation, and population density show a lot of variation across a 64$\,$km$\,\times\,$64$\,$km region. Physical quantities such as temperatures and precipitation are smoother across the region. The two last columns display the fire locations at times $t$ and $t + 1$~\textit{day}, where `no fire' is indicated in gray, `fire' in red, and missing data in black.

The resulting data set contains 18,545 samples. In 58\% of these samples (10,798 samples), the fire increases in size from $t$ to $t + 1$~\textit{day}. In 39\% (7,191 samples), the fire decreases in size. In the remaining samples, the fire stays the same size.

\section{Data Preprocessing}
\label{sec:data_processing}

For the machine learning task, we treat the variables and the `previous fire mask' at time $t$ as data features and the `fire mask' at time $t + 1$~\textit{day} as labels. The values of each data feature, except for the fire masks, are first clipped between a minimum and a maximum clip value, with different clipping values for each feature. We then normalize each feature separately by subtracting the mean and dividing by the standard deviation.

We clip the data because analysis of the features revealed the presence of extreme values, some of which were not even physically reasonable. Moreover, physical variables that span an extensive dynamic range can lead to vanishing or exploding gradients in the deep learning training process. The clipping values were either based on physical knowledge (e.g. percentages between 0-100\%) or set to the  0.1\% and 99.9\% percentiles for each feature. Statistics for processing the input data were calculated over the training data set. Means and standard deviations were calculated after clipping. This allows us to process the data for inference without knowing whether there is currently a fire. 

In our data set, the fire events are generally centered within each  64$\,$km$\,\times\,$64$\,$km region. Therefore, we perform data augmentation in the machine learning input pipeline to offer data examples with fires occurring at different locations. We do so by randomly cropping 32$\,$km$\,\times\,$32$\,$km regions from the original 64$\,$km$\,\times\,$64$\,$km regions.

\section{Machine learning Application}
\label{sec:model}

We illustrate the usefulness of the ‘Next Day Wildfire Spread’ data set by training a deep learning model for fire spread prediction. 

\subsection{Models for segmentation}

We frame the ML task as an image segmentation problem where we classify each area as either containing fire or no fire given the location of the fire on the previous day and the data features described in Section \ref{sec:data}.

\begin{figure}
\centering
\includegraphics[height=0.11\textheight,clip,trim={0.7cm 1.2cm 1.2cm 1cm}]{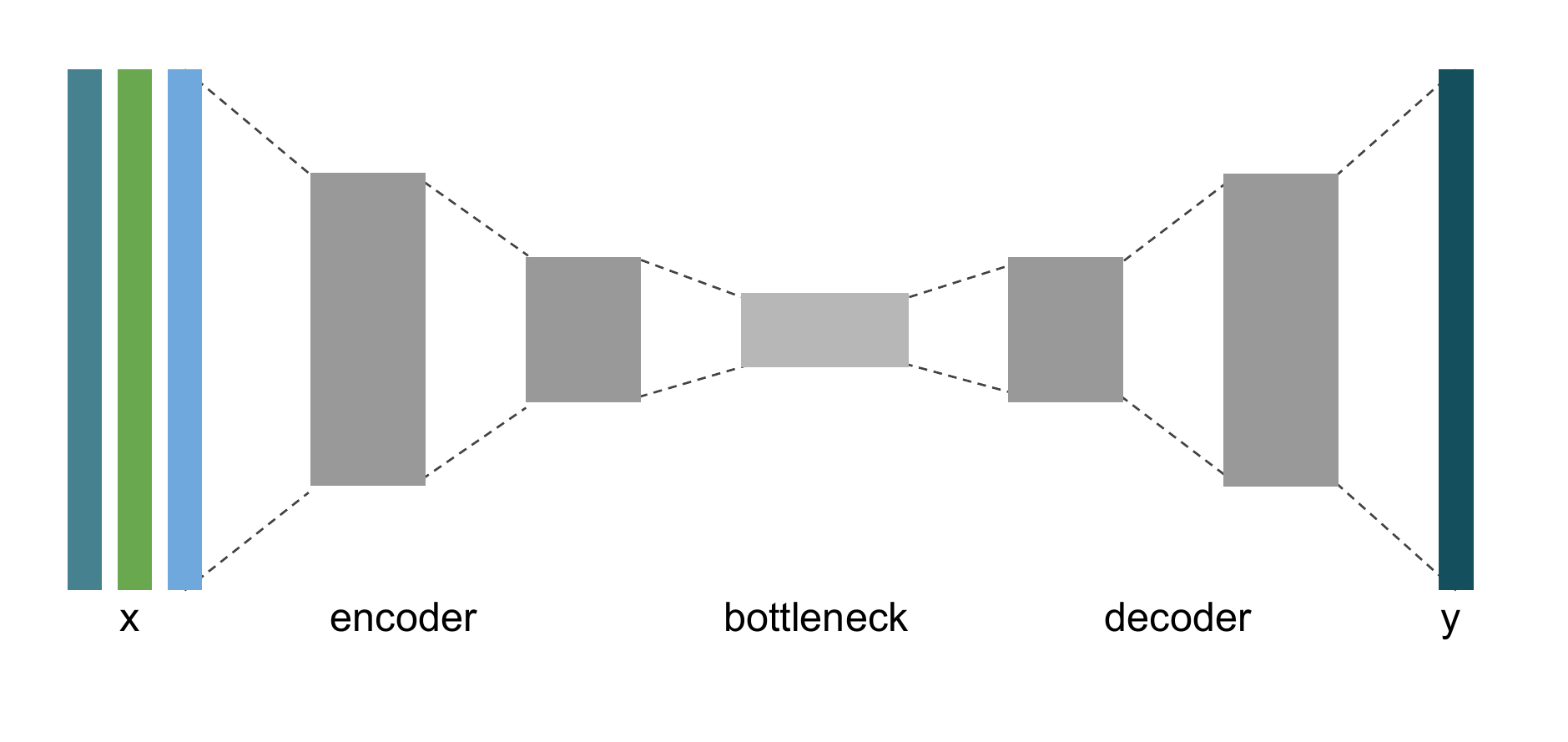}
\caption{Convolutional autoencoder for estimating fire spreading}
\label{fig:ml_models}
\end{figure}

We implement a deep learning model that takes advantage of the spatial information in the input data to perform this task. We use a convolutional autoencoder, \add{a specialized type of neural network for precise image segmentation} 
(\cref{fig:ml_models}). This model was selected because the remote sensing data can be treated as a multi-channel input image and the fire mask as a segmentation map.

In addition to the deep learning model, we also train two machine learning models for comparison, a logistic regression model and a random forest.  These non-deep learning models do not take advantage of the 2D information the way the convolutional autoencoder does. Therefore, with these models, the image segmentation is performed area-wise (pixel-wise). In this setting, the target label for each training sample is a 1$\,$km$\,\times\,$1$\,$km area (pixel). The label is the presence of fire at that area at time $t+1$~\textit{day}. The input variables are all the features for that area at time $t$ and the eight neighboring areas (pixels) around it.

Additional details on model implementation are provided in Appendix \ref{appendix:models}.

\subsection{Training details and hyperparameter tuning}

Since wildfires represent only a small area within a region, we use a weighted cross-entropy loss \cite{aurelio2019learning} and explore a range of different weights on the `fire' labels to \add{take into account} the class imbalance.  Uncertain labels are ignored in the loss and performance calculations. 

For the deep learning model, we include data augmentation by random crop, random flip, and random rotation. We train it over 1000 epochs, with 1000 iteration steps per epoch, with Adam optimizer, on 4 V100 GPUs. We perform the hyperparameter selection by grid search. The resulting network architecture and hyperparameter tuning details are provided in Appendix \ref{appendix:deep_learning}.

The training and hyperparameter details of the logistic regression model and the random forest are described in Appendix \ref{appendix:machine_learning}. 

The best model is selected as the one with the best AUC PR (area under the precision-recall curve). \add{We use AUC PR instead of accuracy because AUC PR is a more effective diagnostic metric for imbalanced binary classification} \cite{sofaer2019area} \add{and our data set contains more `no fire'  than `fire' samples.}

\section{Results}
\label{sec:ongoing}

We compare the  predictions on where the fire will be at time $t + 1$~\textit{day} given data for a region with an ongoing fire at time $t$.

\subsection{Prediction Results}

Metrics on fire spread prediction per area (1$\,$km$\,\times\,$1$\,$km) are shown in Table \ref{table:propagation_segmentation}. All trained models achieve higher AUC than the persistence baseline, that has an AUC of 11.5\%. The model with the highest AUC is the neural network at 28.4\%, followed by the random forest and then logistic regression. The precision and recall for the neural network on the positive class are 33.6\% and 43.1\%, respectively.  The AUC of the logistic regression and random forest models - the non-deep learning models - are within 3\% of one another and as least 6\% lower than the neural network. The logistic regression baseline achieves nearly the same precision as the neural network at 32.5\% but has a lower recall at 35.3\%. The random forest baseline has a higher recall than the neural network at 46.9\% but has a lower precision at 26.3\%.


\begin{table}[!t]
\renewcommand{\arraystretch}{1.3}
\caption{Wildfire spreading prediction metrics}
\label{table:propagation_segmentation}
\centering
\begin{tabular}{l|ccc}
    \hline
    & AUC (PR) & Precision & Recall \\
    \hline
    Neural Network &  28.4 & 33.6 & 43.1 \\
    Random Forest  & 22.5 & 26.3 & 46.9 \\
    Logistic Regression   & 19.8 & 32.5 & 35.3 \\
    Persistence   & 11.5 & 35.7 & 27.3 \\
    \hline
\end{tabular}
\end{table}

\begin{figure}[!t]
  \centering
  \includegraphics[width=\linewidth,clip,trim={0cm 0cm 0cm 0cm}]{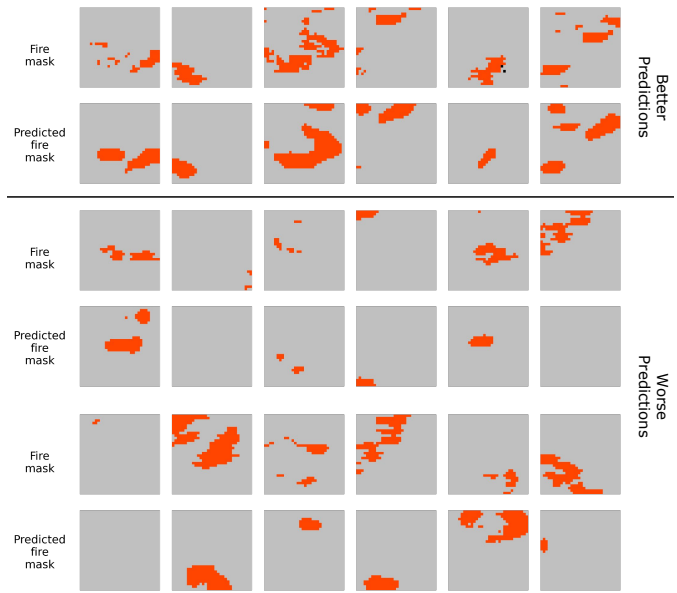}
  \caption{Examples of target labels for next day fire spread and predictions from the deep learning model. Red corresponds to `fire', gray corresponds to `no fire', and black indicates missing data.}
  \label{fig:propagation_segmentation}
\end{figure}

While the metrics on the positive class seem low, visualizations of some samples with predictions and targets in \cref{fig:propagation_segmentation} show that fires are predicted. The predicted fires are roughly in the target location, and are often rounder with smoother borders than the target. Predictions may connect fires that are close together into a single fire. There is a strong dependence on the previous fire mask, with the model predicting that a previous small fire grows while in the target, the fire was no longer present. The segmentation errors result from missing small fires, and misclassified pixels at the boundary between fire and non-fire and between nearby fire areas. 

\begin{figure}[!t]
  \centering
  \includegraphics[width=\linewidth,clip,trim={0cm 0cm 0cm 0cm}]{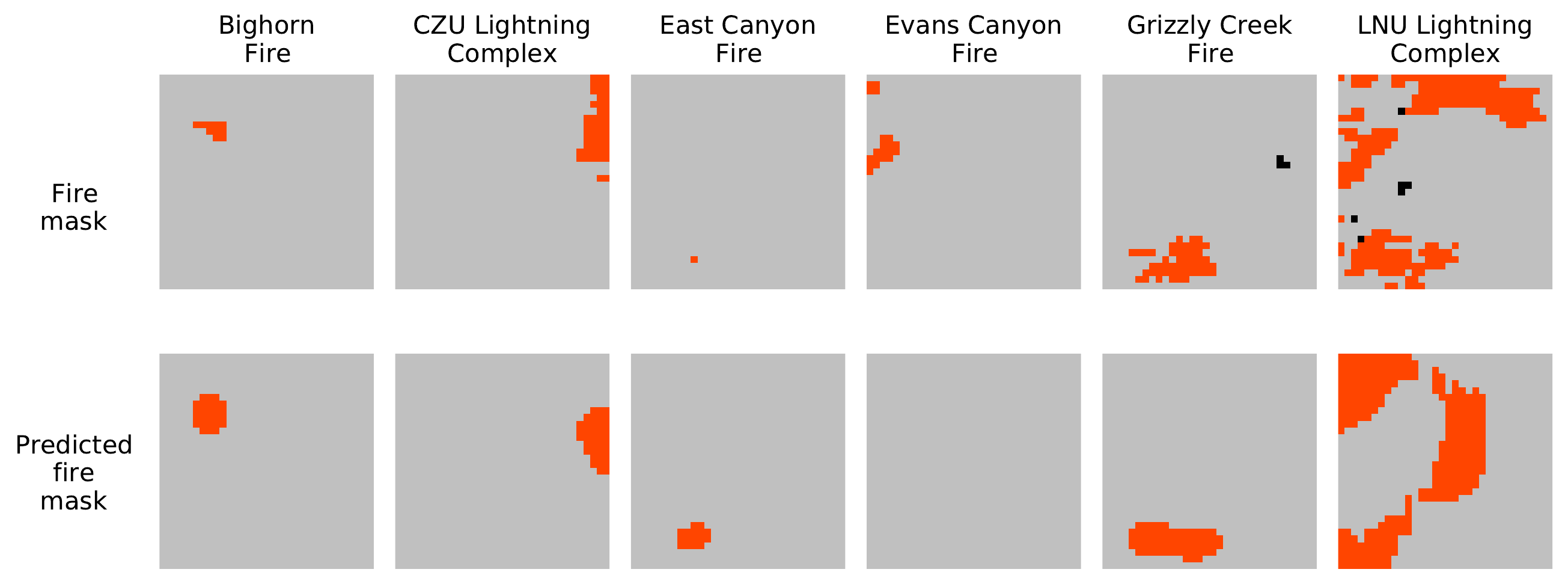}
  \caption{Examples of historical fires with the target labels for next day fire spread and predictions from the deep learning model. The included fires are the Bighorn fire (Santa Catalina Mountains, AZ; 2020-06-07), CZU lightning complex (Santa Cruz Mountains, CA; 2020-08-19), East Canyon fire (Montezuma County, CO; 2020-06-18), Evans Canyon fire (Yakima, WA; 2020-09-02), Grizzly Creek fire (Glenwood Canyon, CO; 2020-08-12), and LNU lightning complex (Sonoma, CA; 2020-08-19). Red corresponds to ‘fire’, gray corresponds to ‘no fire’, and black indicates missing data.}
  \label{fig:historical_fires}
\end{figure}

\add{To illustrate how to use the ML workflow in practice, we evaluate predictions from the deep learning model on a set of six known historical fires. Using our GEE code, we aggregate the input variables at the start of each event and predict the next day's fire spreading. Examples of these known fires are shown in} \cref{fig:historical_fires}. \add{The model achieves an AUC of 38.7\% on these historical fires, higher than that of the test set. The precision of 33.4\% and recall of 37.2\% are similar to the test set metrics. The visualized results are similar to other examples in the data set, with good recognition of larger fires (such as the Grizzly Creek fire). While the exact outline of the fires is not accurate, the model tends to capture the extent of the fire spreading well (for instance, predicting a large fire for the LNU Lightning Complex). The limitations are similar to those observed on the test set: the model tends to predict smoother boundaries than the target (such as the Bighorn fire), can miss smaller fires entirely (such as the Evans Canyon fire), and can merge multiple separated segments of the fire (such as the LNU Lightning Complex).}

\begin{figure}[!t]
  \centering
  \includegraphics[width=\linewidth,clip,trim={0cm 0cm 0cm 0cm}]{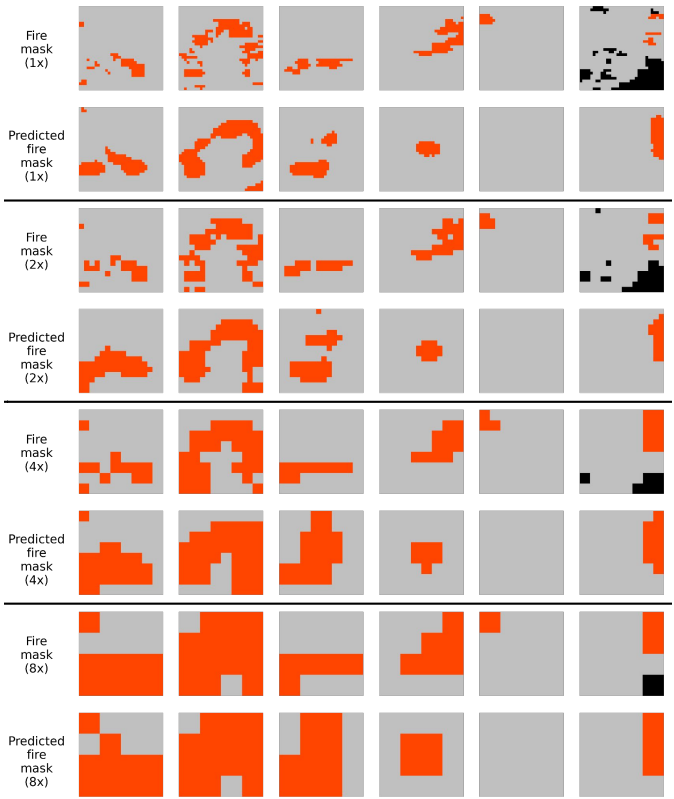}
  \caption{Examples of target labels for next day fire spread and predictions from a deep learning model, where the resolution of the output is lower than the resolution of the input, with the factor indicated in parenthesis. Each column corresponds to a single sample at different target resolutions. Red corresponds to fire, gray corresponds to no-fire, and black indicates missing data.}
  \label{fig:propagation_coarsening}
\end{figure}

Noticing that it is easier to predict the presence of absence of fire over a larger area, we experiment with predicting fire spread at a coarser, lower resolution than the input data. This is to explore and quantify the tradeoff of prediction performance with resolution. That is, predictions for a 2$\,$km$\,\times\,$2$\,$km (2$\,\times\,$2 pixel) area instead of a 1$\,$km$\,\times\,$1$\,$km (1$\,\times\,$1 pixel) area. This turns the problem framing from pixel classification and segmentation into classification over a larger and larger area, as shown in Fig. \ref{fig:propagation_coarsening}. Each row illustrates the target and prediction for fire spread at time $t + 1$~\textit{day} at progressively lower resolution. An area (pixel) within the region is labeled as fire if there is any fire within that area. Similar to previous experiments, the model predicts fire over a larger area than the target. The predictions connect several small areas with fire into a larger fire. The boundaries of the predictions are also smoother than the target. The model sometimes misses lone small fires.

The best precision/recall trade-off for different resolutions is when the output corresponds to an area that is eight times larger than the input area. This results in an AUC of 66.3\%. While the prediction metrics improve as the prediction region size increases, the predictions become less useful from an operational perspective due to less localization of the fire. When predicting at lower spatial resolutions, smaller fires are missed due to the coarser resolution. We experiment with two methods of combining labels for a larger region: a decimal representing the percent of pixels that are on fire, and a binary value that represents any pixel within that region being on fire. For each of these types of labels, experiments are conducted at different resolutions, shown in Table \ref{table:propagation_lowerres}. The neural network performs with higher AUC than the persistence baseline, with most of the gains due to achieving higher precision than the persistence baseline.

\begin{table}[!t]
\renewcommand{\arraystretch}{1.3}
\caption{Lower Resolution Predictions}
\label{table:propagation_lowerres}
\centering
\begin{tabular}{l|cccc}
    \hline
    \multirow{2}{*}{Label} & Lower & \multirow{2}{*}{AUC (PR)} & \multirow{2}{*}{Precision} & \multirow{2}{*}{Recall} \\
                           & Resolution &  &  &  \\
    \hline
    \multirow{3}{*}{Decimal} & 2x &  38.8 & 48.4 & 40.8 \\
    & 4x  & 52.3 & 69.1 & 39.8 \\
    & 8x  & 64.5 & 88.7 & 24.7 \\
    \hline
    \multirow{3}{*}{Binary} & 2x  & 39.8 & 40.3 & 51.9 \\
    & 4x  & 53.5 & 48.8 & 61.0 \\
    & 8x  & 66.3 & 62.1 & 63.1 \\
    \hline
    \multirow{3}{*}{Persistence} & 2x  & 19.5 & 45.6 & 36.8 \\
    & 4x  & 30.6 & 56.0 & 46.8 \\
    & 8x  & 41.3 & 64.0 & 52.5 \\
    \hline
\end{tabular}
\end{table}

\subsection{Feature Analysis}

To analyze the effect of different input features on prediction results, we conduct ablation studies using the model with the highest AUC - the deep learning model. \add{The feature analysis on the random forest and the logistic regression model are provided in Appendices} \ref{appendix:importances} \add{and} \ref{appendix:coefficients}. \add{We do not explicitly program the deep learning model to weigh certain variables more than others. Hence, ablation experiments provide insights into which variables provide the most significant contribution to the model's predictive capabilities.}

Firstly, for each of the input features, we remove that input feature and retrain on the remaining ones. The metrics of the resulting models are summarized in Table \ref{tab:propogation-remove-one}. Removing the current fire location results in the most significant decrease in AUC compared to using all the input features. This \add{result makes intuitive sense and} illustrates the importance of the current fire location as a predictor variable. Among the other input features, removing any one of them results in a similar overall performance to using all the input features, which seems reasonable because many of these features are correlated. Removing some of these input features even resulted in slightly better performance. This increase could be due to randomness, as we have observed that our experiments have a 0.5\% variation in AUC due to random cropping. Additionally, reducing the number of features makes the model simpler, making it less prone to overfitting and reducing the noisiness and redundancies of all the features. 

\begin{table}[!t]
\renewcommand{\arraystretch}{1.3}
\caption{Feature Ablation: Remove one Feature}
\label{tab:propogation-remove-one}
\centering
\begin{tabular}{l|ccc}
    \hline
    Removed feature & AUC (PR) & Precision & Recall \\
    \hline
    Previous fire mask  & 6.8 & 4.6 & 0.2 \\
    Humidity  &  26.1 & 35.4 & 35.7 \\
    Max temperature  &  26.8 & 35.4 & 36.6 \\
    Wind speed  &  27.2 & 33.1 & 42.4 \\
    Vegetation  &  27.7 & 35.0 & 40.0 \\
    Precipitation  &  28.0 & 32.5 & 42.0 \\
    Population  &  28.1 & 32.8 & 43.0 \\
    Wind direction  &  28.3 & 32.8 & 43.1 \\
    Min temperature  &  28.6 & 35.2 & 40.9 \\
    Elevation &  28.8 & 33.6 & 43.0\\
    Drought  &  28.8 & 36.8 & 39.0  \\
    ERC  &  28.8 & 33.0 & 43.2 \\
    \hline
\end{tabular}
\end{table}

Due to the previous fire mask overwhelmingly contributing to the prediction performance, we also retrain the model keeping only two features, the previous fire mask and each of the other features (Table \ref{tab:propogation-keep-two}). We see that vegetation and elevation result in the best performance. The relative ordering of the features changed between leaving a feature out compared to leaving that feature out along with the previous fire.
Because the input features are correlated, we also retrain with only a single feature. Unfortunately, using a single feature resulted in a recall of zero, for all features except for the previous fire mask. Using only the previous fire mask as input results in a recall of 49.6\%, which is higher than the recall of using all features. However, the precision and AUC when only using the previous fire mask is lower, at 29.2\% and 22.9\% respectively.

\begin{table}[!t]
\renewcommand{\arraystretch}{1.3}
\caption{Feature Ablation: Keep Previous Fire Mask and One Other Feature} 
\label{tab:propogation-keep-two}
\centering
\begin{tabular}{l|ccc}
\hline
    Kept feature & AUC (PR) & Precision & Recall \\
    \hline
    Vegetation  &  28.2 & 33.0 & 41.4 \\
    Elevation &  27.0 & 31.3 & 44.4 \\
    Max temperature  &  26.7 & 32.7 & 43.5 \\
    Population  &  26.6 & 32.6 & 43.2 \\
    Precipitation  &  26.6 & 34.1 & 40.6 \\
    Min temperature  &  26.2 & 29.9 & 47.2 \\
    Drought  &  26.0 & 30.8 & 44.1  \\
    Wind speed  &  25.9 & 30.3 & 45.6 \\
    ERC  &  25.8 & 30.7 & 45.0 \\
    Humidity  &  25.7 & 32.4 & 41.7 \\
    Wind direction  &  25.5 & 30.9 & 43.5 \\
\hline
\end{tabular}
\end{table}

\section{\label{SEC_CONCL}Conclusions}

We present this 'Next Day Wildfire Spread' data set as an open data resource for further research to advance our collective ability to anticipate and respond to wildfires. \add{Open data sets allow benchmarking and ML model comparisons, making them essential to producing high-quality ML models for real-world fire scenarios. We create this data set by aggregating nearly a decade of remote-sensing data, combining features including topography, weather, drought index, vegetation, and population density with historical fire records. Using this data set, we demonstrate the potential of deep learning approaches to predict wildfires from remote-sensing data and illustrate the performance gaps on the example task of day-ahead fire spread prediction. Once the ML model is trained, we 
apply it to examples of historical fires, illustrating how it could be used for anticipating the extent of fires the following day. This information could be valuable for allocating resources for fire suppression efforts. 

However, this experiment has some data limitations that must be addressed before incorporating such data-driven approaches into wildfire warning and prediction technologies. The 1 km spatial and daily temporal resolution of the MODIS data limits the ML model's prediction resolution. As such, it does not provide the fine-scale information required for tactical decision-making. As for the GRIDMET weather data, their spatial and temporal resolution is insufficient to capture the local wind patterns that drive fire spreading. 

Since we use the MODIS data for labeling the active fires, our data set comes with the same caveats as the MODIS data. For instance, MODIS does not sample the late afternoon when conditions are most favorable for fire intensification. In addition, it does not separate wildfire events from other prescribed fires. To the extent that prescribed fires correlate with land use, we include population density as a proxy for anthropogenic activity. Moreover, since we focus on the fire spreading pattern, our data aggregation does not include the small fires that occur only on one day, excluding some of the noisy labels from MODIS. Still, we do not explicitly separate wildfires from prescribed fires. The MODIS fire detection algorithm is also conservative, especially at night or when there is cloud coverage. We include an uncertain label that the ML model ignores to account for these pixels, but artificial patterns in fire-spreading labels can still occur. Last but not least, the MODIS labels are affected by fire suppression efforts. While including the population density map can account for this effect to some degree, this is bound to influence the ML model's predictive capabilities. 

These limitations highlight some of the challenges of wildfire prediction from remote-sensing data. Therefore, we release this data set with its GEE data aggregation code, making it more than just a static dataset but an end-to-end framework for aggregating ML data sets from GEE. We designed this code in a modular fashion to easily change the time, the data sources, the geographical area, or the temporal and spatial sampling. As more curated data sources become available on the GEE platform, our framework can be used to improve the data set.} Future data sets could be expanded to a global scale or cover a more extensive time period. Data from recent years, with more fires and larger burnt areas, could be weighted more heavily than data from a decade ago. We could also complement the study with synthetically generated data from high-fidelity fire simulations. Beyond wildfires, the described workflow and methodology could be expanded to other problems such as estimating the likelihood of regions to droughts, hurricanes, and other phenomena from historical remote-sensing data.

\appendices


\renewcommand\thefigure{\thesection-\arabic{figure}}
\renewcommand\thetable{\thesection-\arabic{table}}
\section{Machine Learning Model Details}
\label{appendix:models}
\setcounter{figure}{0}
\setcounter{table}{0}

\subsection{Deep Learning Model}
\label{appendix:deep_learning}

\begin{figure}[h!]
  \centering
  \includegraphics[width=\linewidth,clip,trim={0cm 1cm 0cm 0cm}]{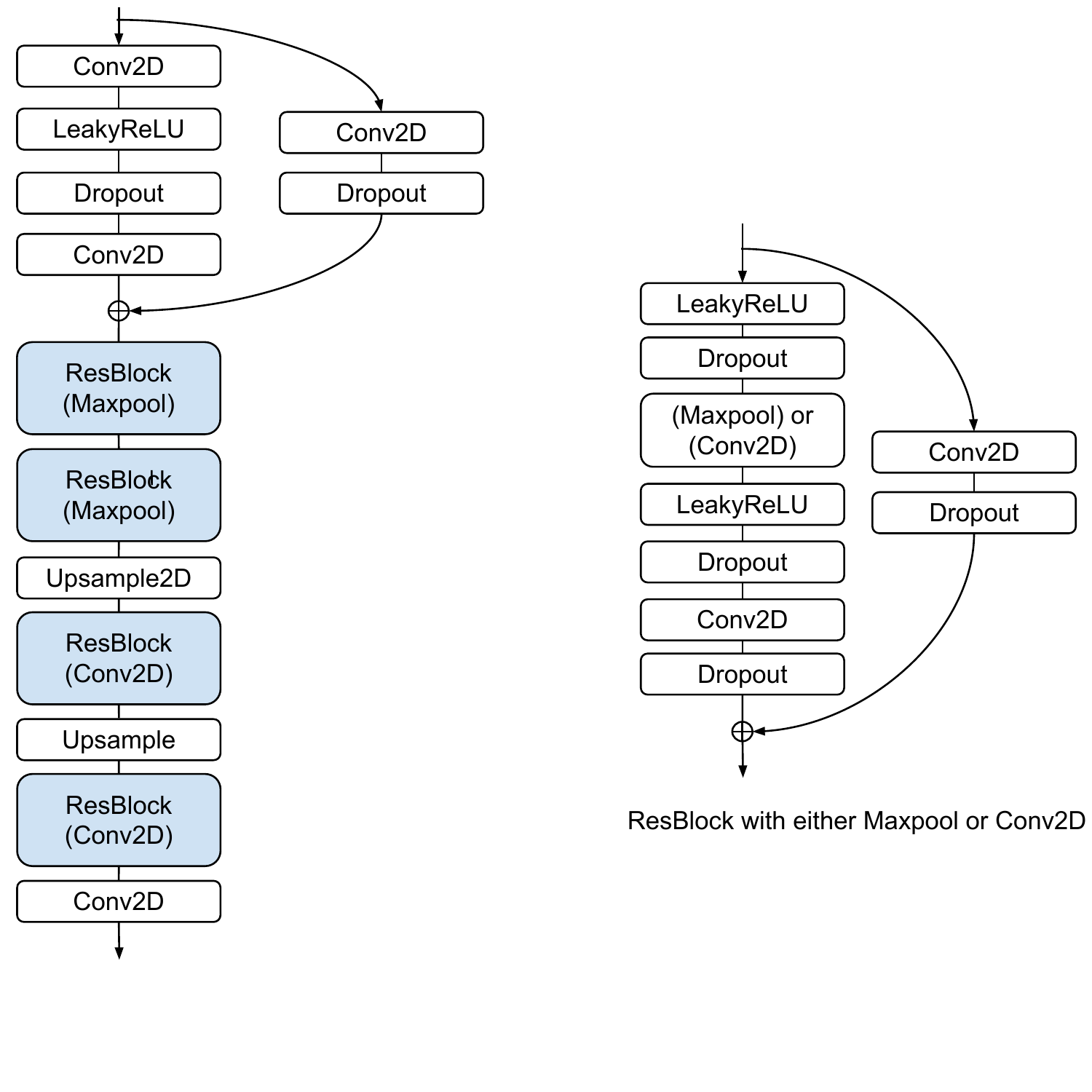}
  \caption{(Left) Deep learning model architecture for image segmentation. The architecture uses a residual block (ResBlock) module with the variation of the ResBlock indicated in parenthesis. (Right) The ResBlock architecture.}
  \label{fig:autoencoder_detailed}
\end{figure}

The architecture of the deep learning model used for image segmentation is shown in Figure \ref{fig:autoencoder_detailed}. All convolutions are 3~$\times$ 3 with a stride of 1 $\times$ 1, pooling is 2 $\times$ 2, and dropout rate is 0.1. All convolutional blocks have 16 filters except for the middle two residual blocks (ResBlocks), which use 32 filters, and the last layer has 1 filter for the segmentation output. The segmentation model is trained with a learning rate of 0.0001 and a weight of 3 on the `fire' class. It is implemented using TensorFlow~\cite{abadi2016tensorflow}.

We perform the hyperparameter selection for the deep learning model by grid search. The best model is selected as the one with the best AUC PR on the validation data set. We implement the number of layers and the number of filters in each layer as hyperparameters. We explore the number of filters in the first convolutional block between 16, 32, and 64. We allow the number of residual blocks in the encoder portion of the network to vary between 1 and 4, with the number of filters doubling in each subsequent block. We define the number of filters in the decoder portion of the segmentation model symmetrically. 

We explore batch sizes 32, 64, 128, 256, learning rates 0.01, 0.001, 0.0001, and 0.00001, dropout rates in increments of 0.1, and adding L1 and L2 regularization in powers of 10.

\subsection{Non-Deep Learning Models}
\label{appendix:machine_learning}

The two non-deep learning models are implemented using Scikit-learn \cite{scikit-learn}. For these models, the training data set is balanced to contain fire and non-fire examples in an approximately 1:1 ratio. Metrics are calculated on the entire test data set with no balancing. Unknown labels are ignored during training and evaluation.

With these models, the image segmentation is performed area-wise (pixel-wise). Consequently, the input variables correspond to a single area (pixel) in the input region.  To provide spatial context, we include the eight neighboring areas (pixels) in a square around the input area (pixel). To avoid missing data, we exclude the 1-pixel border around the image during training and evaluation.

For the logistic regression model, we use the default hyperparameters from the Scikit-learn package. For the random forest, we sweep over values of the maximum tree depth between 1 and 20. As we increase the maximum depth, precision increases while recall decreases. Therefore, we select the value with the highest AUC (PR) on the validation set. Using this methodology, we select a maximum depth of 15. All other hyperparameters are set to the default values from the Scikit-learn package.

Initial results from both models show that they over-predict the presence of fire, with very low precision and high recall. Therefore, we weigh the `no fire' class using a weight defined as follows:
\[ 
\frac{\textrm{number of examples}}{\textrm{number of `no fire' examples}} \times W,
\]
where $W$ is a factor we adjust to increase the weight. We search over increasing values of $W$ from 1 to 8. We select the value of the $W$ that yields the highest AUC (PR) on the validation data set. For logistic regression, $W$ is 7, and for the random forest, $W$ is 5.

\subsection{Random Forest Feature Analysis}
\label{appendix:importances}

The 20 most important features in the random forest model are reported in Table \ref{table:trees_importance}. Feature importance is considered as the mean decrease in Gini impurity brought by splitting on that feature across the trees. The Gini impurity of a set of points after a split is the probability of misclassification if each point is randomly classified according to the distribution of labels in the set of points. Features are numbered in row-major order for the nine pixels (3 $\times$ 3 square) included in each example. Pixel number 5 is the input pixel. For example, `previous fire mask (3)' is the previous fire mask of the top-right neighbor of the input pixel. Based on this analysis, the previous fire mask of the input pixel is the most important feature, followed by the previous fire mask of the neighboring pixels. ERC follows as the next most important feature, and the remaining features all have very low importance. Precipitation for all pixels is the least important feature.

\begin{table}[!h]
\renewcommand{\arraystretch}{1.3}
\caption{Random Forest: Gini Importance}
\label{table:trees_importance}
\centering
\begin{tabular}{l|c}
    \hline
    Feature & Gini Importance \\
    \hline
    Previous fire mask (5)  &  0.092 \\
    Previous fire mask (6)  &  0.074 \\
    Previous fire mask (4)  &  0.073 \\
    Previous fire mask (2)  &  0.071 \\
    Previous fire mask (1)  &  0.068 \\
    Previous fire mask (8)  &  0.067 \\
    Previous fire mask (9)  &  0.061 \\
    Previous fire mask (3)  &  0.052 \\
    Previous fire mask (7)  &  0.051 \\
    ERC (7)  &  0.017 \\
    ERC (8)  &  0.014 \\
    ERC (5)  &  0.014 \\
    ERC (6)  &  0.012 \\
    ERC (9)  &  0.01 \\
    ERC (2)  &  0.008 \\
    ERC (1)  &  0.008 \\
    ERC (4)  &  0.008 \\
    ERC (3)  &  0.007 \\
    Elevation (7)  &  0.007 \\
    Elevation (9)  &  0.005 \\
    \hline
\end{tabular}
\end{table}

\subsection{Logistic Regression Feature Analysis}
\label{appendix:coefficients}

The 20 largest magnitude coefficients of the logistic regression model by absolute value are reported in Table \ref{table:logistic_coefficients}. Features are numbered in row-major order for the nine pixels (3 $\times$ 3 square) included in each example. Pixel number 5 is the input pixel. For example, "Min temp. (3)" is the minimum temperature of the top-right neighbor of the input pixel.

We conduct a feature ablation study by comparing the results of removing each feature during training. The resulting metrics are reported in Table \ref{table:logistic_ablation}. This analysis demonstrates that the previous fire mask is the most important feature for the logistic regression model, similar to the other models.

\begin{table}[!h]
\renewcommand{\arraystretch}{1.3}
\caption{Logistic Regression: Coefficients}
\label{table:logistic_coefficients}
\centering
\begin{tabular}{l|ccc}
    \hline
    \multirow{2}{*}{Feature} & \multirow{2}{*}{Coefficient} & Standard & \multirow{2}{*}{Units} \\
    & & Deviation & \\
    \hline
    Min temp. (3)  &  -7.2  &  8.98 & K  \\
    Max temp. (3)  &  6.01  &  9.82 & K  \\
    ERC (1)  &  -5.69  &  20.85 & index  \\
    Max temp. (5)  &  -5.49  &  9.82 & K  \\
    ERC (5)  &  4.97  &  20.85 & index  \\
    Min temp. (5)  &  4.77  &  8.98 & K  \\
    Humidity (1)  &  -3.64  &  0.0 & kg/kg  \\
    Max temp. (8)  &  -3.5  &  9.82 & K  \\
    Min temp. (8)  &  3.24  &  8.98 & K  \\
    Min temp. (4)  &  -3.16  &  8.98 & K  \\
    Max temp. (1)  &  2.87  &  9.82 & K  \\
    ERC (7)  &  2.71  &  20.85 & index  \\
    Precipitation (1)  &  -2.44  &  4.48 & mm  \\
    Precipitation (5)  &  2.43  &  4.48 & mm \\
    Max temp. (2)  &  -2.31  &  9.82 & K  \\
    Precipitation (9)  &  -2.09  &  4.48 & mm  \\
    Min temp. (2)  &  1.96  &  8.98 & K  \\
    Min temp. (1)  &  1.87  &  8.98 & K  \\
    ERC (4)  &  -1.81  &  20.85 & index  \\
    ERC (3)  &  1.73  &  20.85 & index  \\
    \hline
\end{tabular}
\end{table}

\begin{table}[!h]
\renewcommand{\arraystretch}{1.3}
\caption{Logistic Regression: Remove One Feature}
\label{table:logistic_ablation}
\centering
\begin{tabular}{l|ccc}
    \hline
    Removed feature & AUC (PR) & Precision & Recall \\
    \hline
    Previous fire mask  &  3.1  &  0.0  &  0.0  \\
    ERC  &  19.6  &  32.4  &  35.5  \\
    Elevation  &  19.6  &  32.3  &  35.5  \\
    Drought  &  19.6  &  32.4  &  35.4  \\
    Min temperature  &  19.6  &  32.2  &  35.5  \\
    Wind direction  &  19.7  &  32.3  &  35.7  \\
    Max temperature  &  19.7  &  32.3  &  35.5  \\
    Vegetation  &  19.8  &  32.3  &  35.6  \\
    Precipitation  &  19.8  &  32.3  &  35.6  \\
    Population  &  19.8  &  32.3  &  35.6  \\
    Humidity  &  19.8  &  32.4  &  35.6  \\
    Wind speed  &  20.0  &  32.1  &  35.7  \\
    \hline
\end{tabular}
\end{table}

\clearpage
\onecolumn
\section{Additional Visualizations}
\setcounter{figure}{0}

\begin{figure*}[h!]
  \centering
  \includegraphics[width=0.92\linewidth,clip,trim={0cm 0cm 0cm .2cm}]{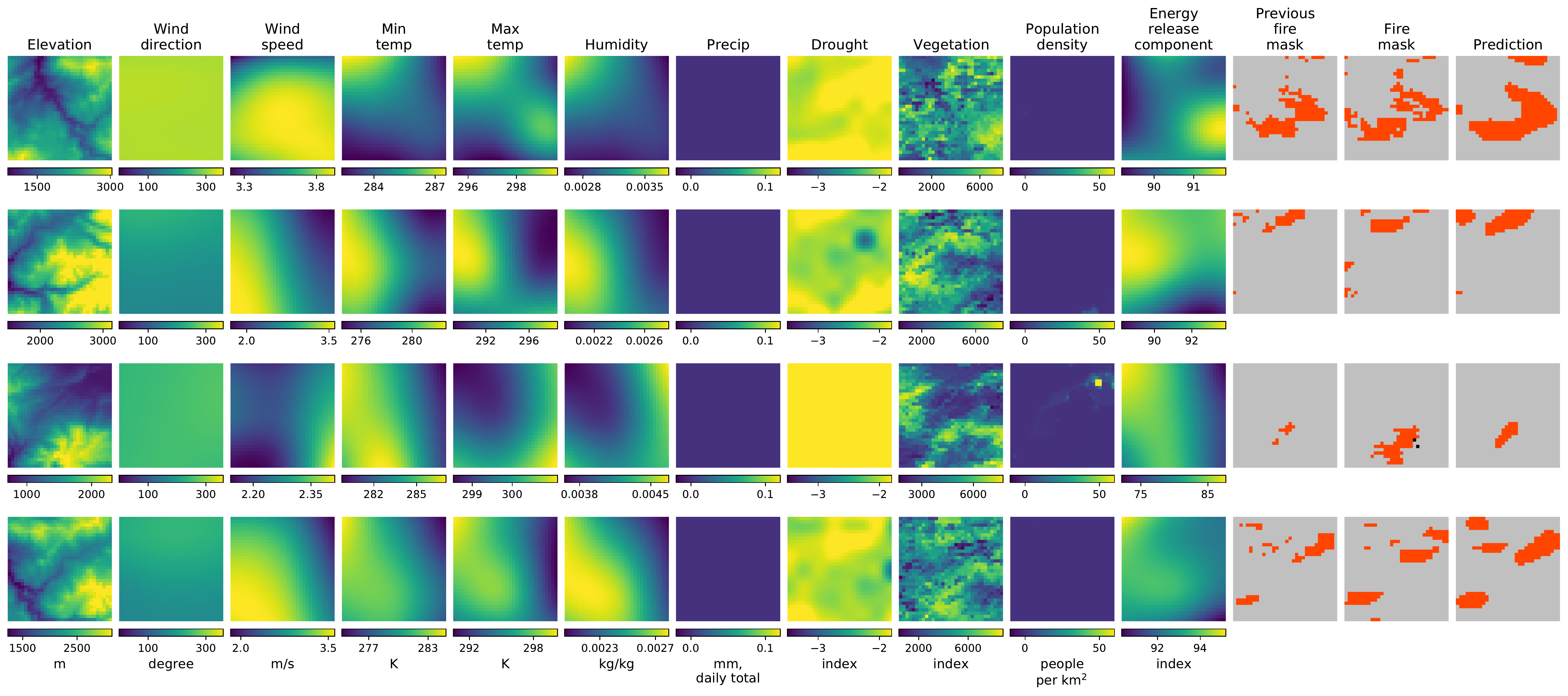}
  \caption{Examples of better segmentation predictions. Each row is a sample.}
  \label{fig:app_segmentation_good}
\end{figure*}

\begin{figure*}[h!]
  \centering
  \includegraphics[width=0.92\linewidth,clip,trim={0cm 0cm 0cm .2cm}]{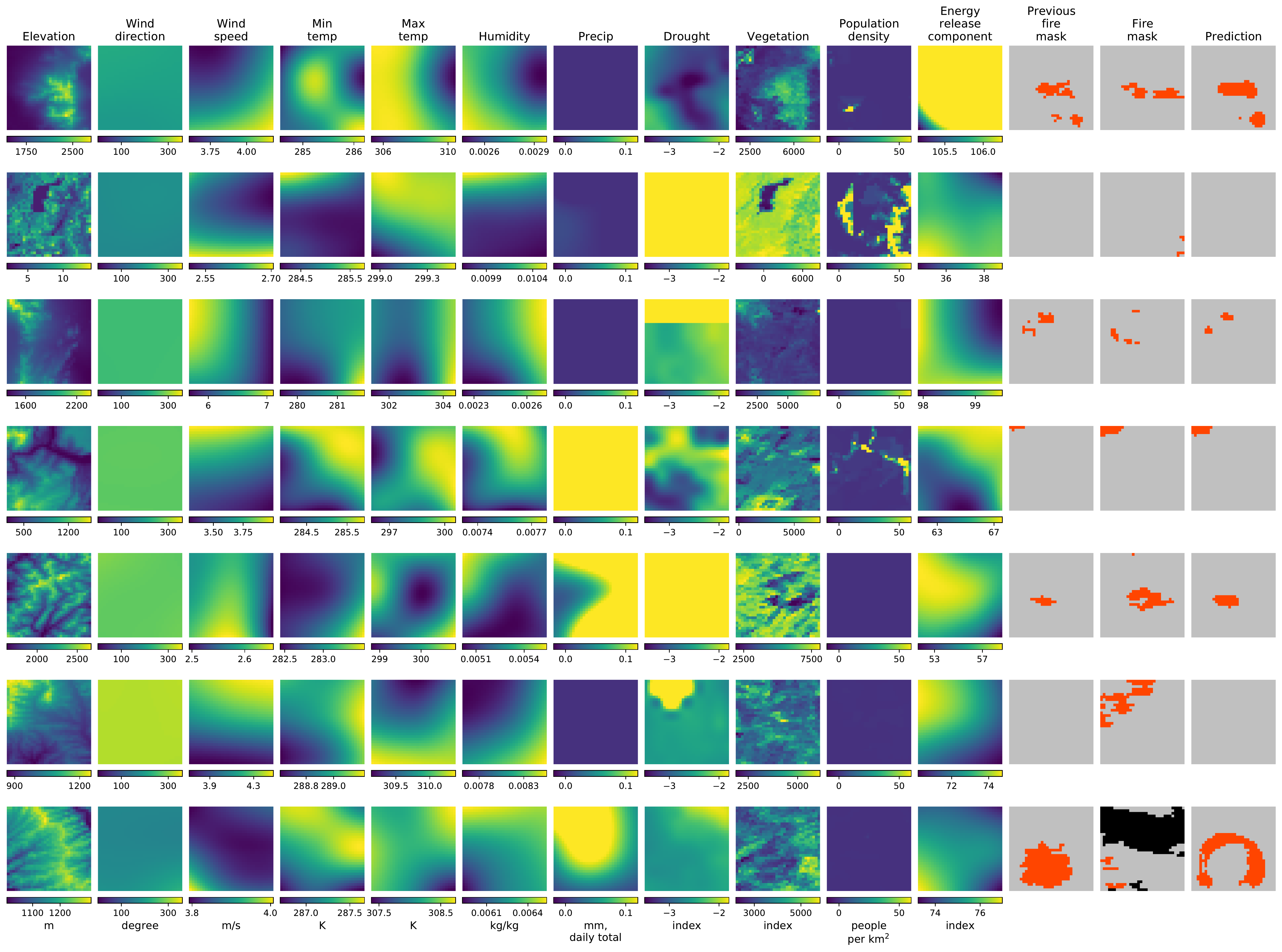}
  \caption{Examples of worse segmentation predictions. Each row is a sample.}
  \label{fig:app_segmentation_bad}
\end{figure*}

\begin{figure*}[h!]
  \centering
  \includegraphics[width=0.7\linewidth,clip,trim={0cm 0cm 0cm 0cm}]{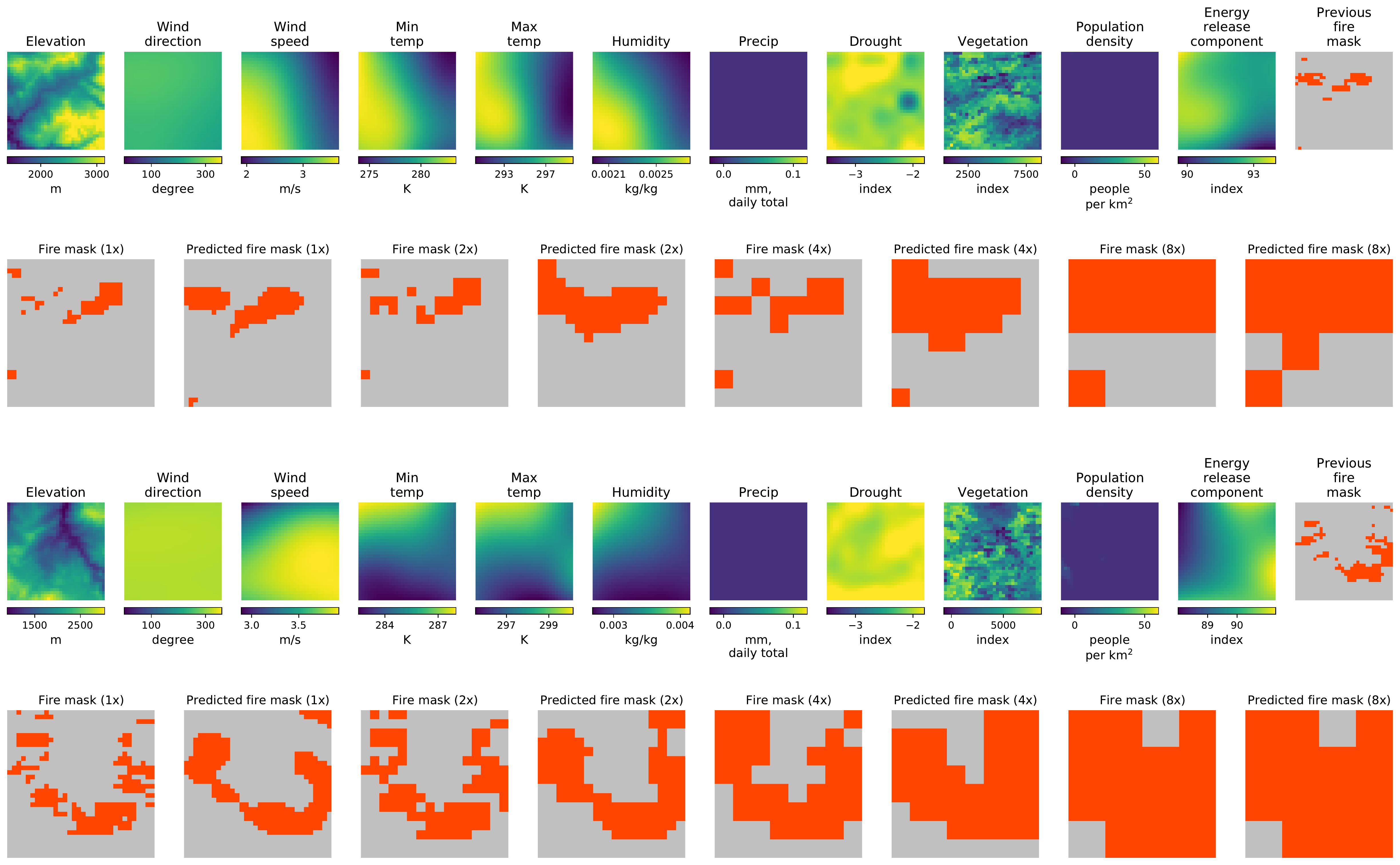}
  \caption{Examples of better coarse segmentation predictions. Each pair of rows is a sample.}
   \label{fig:app_coarse_segmentation_good}
\end{figure*}

\begin{figure*}[h!]
  \centering
  \includegraphics[width=0.7\linewidth,clip,trim={0cm 0cm 0cm 0cm}]{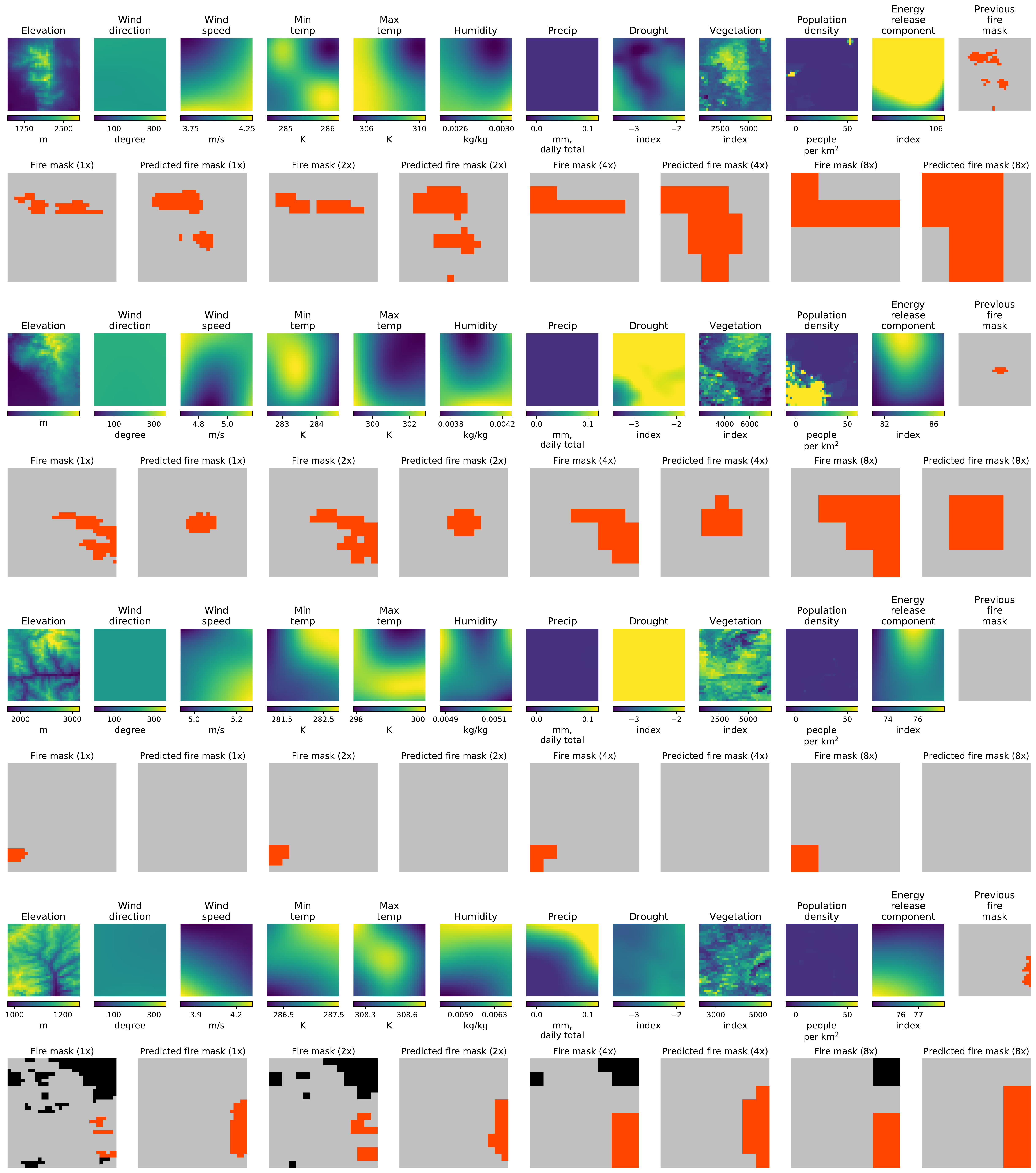}
  \caption{Examples of worse coarse segmentation predictions. Each pair of rows is a sample.}
    \label{fig:app_coarse_segmentation_bad}
\end{figure*}

\Cref{fig:app_segmentation_good} and \cref{fig:app_segmentation_bad} provide additional examples of fire spreading prediction results. \Cref{fig:app_coarse_segmentation_good} and \cref{fig:app_coarse_segmentation_bad} provide additional examples of prediction results at coarser resolution.

\twocolumn
\section*{Acknowledgments}

We would like to thank Zack Ontiveros, Carla Cromberg, Jason Hickey, Qing Wang, Cenk Gazen, Shreya Agrawal, Winston Huang, David Tian, Renaud Couzin, Ian Post, Michael Brenner, Fei Sha, John Burge, Zvika Ben-Haim, Avichai Tendler, Sella Nevo, and Omer Nevo.
We thank the Stanford Exploration Project affiliate members for their financial support.

\clearpage

\doublespacing

\bibliographystyle{IEEEtran}
\bibliography{IEEEabrv,references}
%



%




\singlespacing

\begin{IEEEbiographynophoto}{Fantine Huot}
Fantine Huot is a Ph.D. Candidate in Geophysics at Stanford University and is passionate about tackling computational challenges for physical sciences. She has worked on various topics, including seismology, remote sensing, deep learning, and high-performance computing. She holds an M.S. in Science and Executive Engineering from the Ecole Nationale Superieure des Mines de Paris.
\end{IEEEbiographynophoto}

\begin{IEEEbiographynophoto}{R. Lily Hu}
R. Lily Hu is a researcher and engineer at Google Research. She conducted research in machine learning for the physical sciences and sustainability, drawing upon computer vision, decision analysis, and optimization. She earned a MS/Ph.D in engineering from the University of California, Berkeley and a BASci in engineering science from the University of Toronto.
\end{IEEEbiographynophoto}

\begin{IEEEbiographynophoto}{Nita Goyal}
Nita Goyal has been a software engineer and researcher at Google since 2013. She has worked in various fields including natural language understanding, search and recommendations and machine learning for climate and sustainability. She has previously been in several tech startups and earned a Ph.D. in Computer Science from Stanford University and a B.Tech. in CS from IIT Kanpur, India. She is the first woman CS graduate from any of the IITs.
\end{IEEEbiographynophoto}

\begin{IEEEbiographynophoto}{Tharun Sankar}
Tharun Sankar has been with Google Research as a Software Engineer since 2020 after receiving his B.A. in computer science from Cornell University. His research work has spanned fields such as computer vision, audio processing, computational sustainability, and natural language processing.
\end{IEEEbiographynophoto}

\begin{IEEEbiographynophoto}{Matthias Ihme}
Matthias Ihme is Assistant Professor of Mechanical Engineering at Stanford University. He obtained is Ph.D. in 2008 from Stanford University, and served on the faculty in the Department of Aerospace Engineering at the University of Michigan. His primary research interests are the prediction and modeling of turbulent reacting flows, scalar mixing, and subgrid-scale modeling for large eddy simulation. He is a member of the Combustion Institute, APS, and AIAA, where he serves on the Gas Turbine Engines Technical Committee. He is recipient of the NSF CAREER Award, the AFOSR Young Investigator Award, the ONR Young Investigator Award, and a NASA Early Career Faculty Award.
\end{IEEEbiographynophoto}

\begin{IEEEbiographynophoto}{Yi-fan Chen}
Yi-Fan Chen has been a Software Engineer and Researcher with Google since 2012. He earned his Ph.D in applied physics from Cornell University and has worked in various fields including: nonlinear and ultrafast optics, photolithography, video processing, recommendation systems and high performance computing. 
\end{IEEEbiographynophoto}

\vfill






\end{document}